%% file: main.tex
\documentclass[runningheads]{llncs}

\usepackage{eccv}

\usepackage{eccvabbrv}

\BeforeBeginEnvironment{wrapfigure}{\setlength{\intextsep}{0pt}}
\usepackage{graphicx}
\usepackage{booktabs}
\usepackage{wrapfig}
\usepackage{multirow}
\usepackage[table]{xcolor}
\usepackage[ruled]{algorithm2e} 

\usepackage[accsupp]{axessibility}  

\usepackage{hyperref}

\usepackage{orcidlink}

\begin{document}

\title{SAP: Segment Any 4K Panorama} 

\titlerunning{SAP: Segment Any 4K Panorama}




\author{
Lutao Jiang $^{1*}$ \and
Zidong Cao $^{1*}$ \and
Weikai Chen $^{2\dag}$ \and
Xu Zheng $^{1}$ \and
Yuanhuiyi Lyu $^{1}$ \and
Zhenyang Li $^{4}$ \and
Zeyu Hu $^{2}$ \and
Yingda Yin $^{2}$ \and
Keyang Luo $^{2}$ \and
Runze Zhang $^{2}$ \and
Kai Yan $^{2}$ \and
Shengju Qian $^{2}$ \and
Haidi Fan $^{2}$ \and
Yifan Peng $^{4}$ \and
Xin Wang $^{2}$ \and
Hui Xiong $^{1,3}$ \and
Ying-Cong Chen $^{1,3\dag}$
}

\authorrunning{L. Jiang et al.}

\institute{The Hong Kong University of Science and Technology (Guangzhou) \and
LIGHTSPEED \and
The Hong Kong University of Science and Technology \and
The University of Hong Kong}

\maketitle
\renewcommand{\thefootnote}{} 
\footnotetext{
    $^*$ Co-first author. \hspace{0.8em} $^\dag$ Corresponding author.
}
\vspace{-25pt}
\begin{abstract}
Promptable \textbf{instance segmentation} is widely adopted in embodied and AR systems, yet the performance of foundation models trained on perspective imagery often degrades on 360° panoramas.
In this paper, we introduce \textbf{Segment Any 4K Panorama (SAP)}, a foundation model for 4K high-resolution panoramic instance-level segmentation. 
We reformulate panoramic segmentation as fixed-trajectory perspective video segmentation, decomposing a panorama into overlapping perspective patches sampled along a continuous spherical traversal. 
This memory-aligned reformulation preserves native 4K resolution while restoring the smooth viewpoint transitions required for stable cross-view propagation.
To enable large-scale supervision, we synthesize \textbf{183,440} 4K-resolution panoramic images with instance segmentation labels using the InfiniGen engine.
Trained under this trajectory-aligned paradigm, SAP generalizes effectively to real-world 360° images, achieving \textbf{+17.2 zero-shot mIoU gain} over vanilla SAM2 of different sizes on real-world 4K panorama benchmark. Project page: \href{https://lutao2021.github.io/SAP_Page/}{Project Page}.
\end{abstract}

\vspace{-30pt}
\input{sections/intro}
\input{sections/related}

\input{sections/method}

\input{sections/dataset}

\input{sections/exp}

\input{sections/conclusion}

%
%
\clearpage
\bibliographystyle{splncs04}
\bibliography{main}
\nocite{jiang2024general, jiang2025dimer, hua2025sat2city}

\clearpage
\input{sections/appendix}
\end{document}

%% file: sections/intro.tex
\section{Introduction}
\label{sec:intro}
\vspace{-10pt}

As 360° cameras become increasingly prevalent in robotics, AR/VR, and embodied agents~\cite{xia2018gibson, zheng2025panorama}, there is growing demand for robust high-resolution {instance segmentation} on panoramic imagery. While recent segmentation foundation models~\cite{kirillov2023segment, ravi2024sam, carion2025sam} have demonstrated impressive generalization on perspective views, their performance degrades noticeably on 360° panoramas, which differ in both visual content density and underlying imaging geometry. Furthermore, most existing panoramic segmentation methods~\cite{yang2020omnisupervised, yang2021capturing, zhong2025omnisam, li2023sgat4pass, zheng2023look, yang2020ds, zheng2023both, zhang2024goodsam, zhang2022bending, zheng2024open, zhang2024behind, zheng2024semantics, zheng2024360sfuda++, xu2025mamba4pass, chamseddine2026panosamic} primarily focus on semantic segmentation, and high-resolution {instance-level} panoramic segmentation remains underexplored.

\begin{figure}[t!]
    \centering
    \includegraphics[width=1\linewidth]{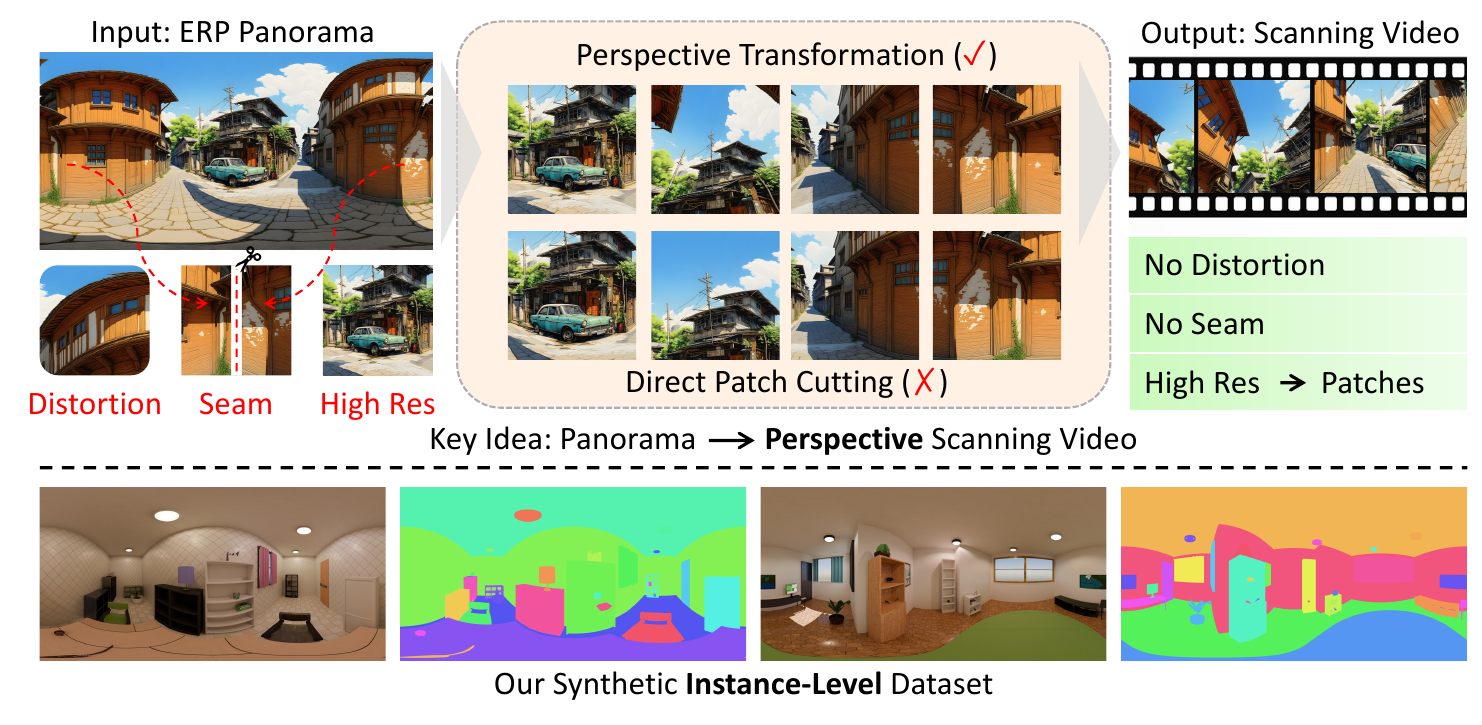}
    \vspace{-20pt}
    \caption{The main challenges of equirectangular panoramic segmentation include (i) severe distortions compared to standard perspective images, (ii) discontinuities at the left-right seam and near the poles, and (iii) ultra-high resolution that is rarely addressed by conventional segmentation methods. Our key idea is to convert a panorama into a fixed-trajectory perspective scanning video to avoid these issues and to provide high-quality synthetic data for fine-tuning segmentation models.}
    \vspace{-10pt}
    \label{fig: teaser}
\end{figure}

As shown in Fig.~\ref{fig: teaser}, panoramic segmentation at native 4K resolution presents both computational and structural challenges. Panoramas contain substantially more visual content than perspective frames and are often resized to fit square-image segmentation models. For instance, SAM-series models support only $1024^2$ inputs. 
A 4K 2:1 panorama with $360^\circ\times180^\circ$ FoV is typically resized to $1024 \times 512$ and padded to $1024 \times 1024$, discarding fine details while wasting computation on padded regions.
In addition, equirectangular projection introduces geometric distortion and artificial seam discontinuities, where the left and right boundaries are adjacent on the sphere but disconnected in the raster representation, frequently leading to fragmented masks and cross-boundary inconsistencies.

OmniSAM~\cite{zhong2025omnisam} proposes to utilize SAM2~\cite{ravi2024sam}, a video segmentation model, to tackle panoramic semantic segmentation by sliding-window cropping the panorama into overlapping patch sequences.
However, we contend that distortion and seam artifacts are surface-level symptoms of a deeper problem: spherical panoramas fundamentally violate the structural assumptions underlying streaming memory models. 
Models such as SAM2~\cite{ravi2024sam} are designed for temporally ordered inputs, where consecutive frames correspond to smooth camera motion and overlapping visual content. 
In contrast, an equirectangular panorama provides no intrinsic notion of temporal order. 
When decomposed through sliding windows, the induced sequence breaks physical viewpoint continuity. 
As a result, the memory mechanism operates under violated assumptions, leading to unstable identity propagation and inconsistent segmentation.

We therefore \textbf{cast panoramic segmentation as a topology–memory alignment problem} with respect to the temporal continuity assumptions of memory-based segmentation models.
As illustrated in Fig.~\ref{fig: teaser}, naïvely resizing or independently processing panoramic patches breaks the temporal continuity assumed by streaming memory models. 
Instead of operating directly in equirectangular space, we convert a 360° panorama into overlapping perspective patches sampled along a fixed trajectory.
This trajectory induces a physically meaningful progression over the sphere, preserving native resolution while enforcing smooth transitions between views. 
As a result, panoramic segmentation can be reformulated as perspective video segmentation under a controlled camera path, restoring compatibility with the memory propagation mechanism in SAM2. 
By operating entirely in perspective projections, our approach aligns panoramic inputs with the inductive biases of perspective-trained foundation models, thereby effectively leveraging their learned priors for robust instance reasoning.

Beyond structural misalignment, supervision is also limited.
Large-scale high-resolution panoramas with \textbf{instance-level} segmentation labels are scarce due to the high annotation cost and the complexity introduced by projection distortions and seam consistency. 
Existing foundation models are predominantly trained on perspective data and lack sufficient supervision for panoramic geometry.
This data gap makes it difficult to systematically adapt foundation models to panoramic inputs. 
To address this issue, we synthesize \textbf{183K} high-resolution 4K panoramic images with instance segmentation labels using the InfiniGen data engine~\cite{raistrick2024infinigen}. This large-scale supervision enables trajectory-based training at the resolution and viewpoint coverage required by panoramic perception.

Based on this reformulation and the collected large-scale dataset, we introduce \textbf{Segment Any 4K Panorama (SAP)}, a promptable instance segmentation foundation model designed for 4K high-resolution panoramic inputs. 
SAP leverages our panoramic training data to explicitly strengthen the segmentation capability under the fixed-trajectory perspective sequences in SAM2. This design targets the failure modes of perspective-trained models on 360° imagery while maintaining compatibility with established segmentation foundations.
Furthermore, to prevent catastrophic forgetting~\cite{kirkpatrick2017overcoming}, we integrated SAM2's original training data to maintain the SAP's versatility in real-world applications.
Despite being trained on synthetic panoramas, SAP transfers effectively to real-world 360° imagery, delivering an average of \textbf{+17.2 zero-shot mIoU gains} over vanilla SAM2 across model sizes. 
These results demonstrate that resolving topology–memory mismatch through trajectory-aligned traversal and large-scale supervision substantially narrows the gap between perspective-trained segmentation foundations and panoramic deployment.

In summary, our contributions are threefold:
\begin{itemize}
    \item We introduce Segment Any 4K Panorama (SAP), a {promptable instance segmentation foundation model} for high-resolution panoramas, yielding +17.2 zero-shot mean mIoU gains over vanilla SAM2 (cross model sizes) on 4K real-world panoramas.
    \item We construct a large-scale dataset containing 183K instance-labeled 4K panoramas, addressing the scarcity of panoramic segmentation data.
    \item We cast panoramic instance segmentation as a topology–memory alignment problem, restoring compatibility between spherical topology and memory-based model through fixed-trajectory perspective video reformulation.

\end{itemize}

%% file: sections/related.tex
\section{Related Works}

\subsection{Panoramic Image Segmentation}

Panoramic images capture a comprehensive 360$^{\circ}$$\times$180$^{\circ}$ field-of-view (FoV) but inherently suffer from severe spherical distortions in the ERP representation~\cite{yang2020ds,yang2020omnisupervised,yang2021capturing,ai2025survey}. To mitigate this, previous methods developed distortion-aware modules, such as Spherical CNNs~\cite{lee2019spherephd, su2017learning,zhang2019orientation} and Transformers~\cite{li2023sgat4pass} based on the spherical transformation to process the distortion. For example, Trans4PASS~\cite{zhang2022bending} utilizes Deformable Patch Embedding (DPE) to learn object deformations in the spherical domain, while PanoFormer~\cite{shen2022panoformer} extracts tangent patches from the spherical domain and uses learnable token flows to reduce distortion effects. The scarcity of pixel-exact annotated panoramic data remains a primary obstacle for training robust panoramic foundation models. Consequently, a large branch of research~\cite{zhang2022bending,zheng2023both,zheng2023look,zhang2024behind,zhong2025omnisam} has been forced to rely on Unsupervised Domain Adaptation (UDA)~\cite{liu2022deep} to transfer knowledge from label-rich pinhole datasets~\cite{cordts2016cityscapes} to unlabeled panoramas. Another direction is to project the ERP image into perspective formats with multiple views, such as tangent patches (TP)~\cite{eder2020tangent}. However, processing multiple perspective views simultaneously incurs significant computational overhead, and independent view processing disrupts instance continuity across projection boundaries~\cite{cao2025st}, leading to fragmented masks. Extending these paradigms to ultra-high resolutions (\textit{e.g.}, 4K) further results in out-of-memory (OOM) failures or requires downsampling that discards small-object details.   
Furthermore, while the majority of previous methods focus on panoramic semantic segmentation, we focus on instance segmentation such as SAM.

\subsection{Segmentation Vision Foundation Models}

The Segment Anything Model (SAM)~\cite{kirillov2023segment} has established a new standard for zero-shot segmentation. However, directly applying SAM to high-resolution images is constrained due to memory bottlenecks. The following SAM2~\cite{ravi2024sam} extends the zero-shot capability to the spatio-temporal domain via a streaming memory mechanism, including a memory encoder, memory bank, and memory attention module. As a result, SAM2 achieves impressive video object segmentation by tracking objects across frames. Recently, OmniSAM~\cite{zhong2025omnisam} feeds continuous ERP sliding windows into SAM2 to enhance cross-patch feature alignment. However, applying sliding windows directly on ERP images forces the model to process severe polar distortions, which conflicts with the pre-trained perspective priors inherent in SAM2. In contrast, our SAP transforms a 4K panoramic image into a sequence of distortion-free perspective frames arranged in a continuous and overlapping trajectory. By feeding such a perspective video into SAM2, the spatial consistency across adjacent perspective views is preserved by utilizing the streaming memory mechanism in SAM2.

%% file: sections/method.tex
\section{Method}

\subsection{Task Definition}
The Segment Anything Model (SAM)~\cite{kirillov2023segment} formulates segmentation as a promptable mask prediction task. 
Given an image $I \in \mathbb{R}^{H \times W \times 3}$ and a prompt point $\mathbf{p} \in \mathbb{R}^{2}$, SAM predicts a binary mask $M \in \{0,1\}^{H \times W}$:
\begin{equation}
M = f_{\theta}(I, \mathbf{p}).
\end{equation}

We extend this formulation to panoramic images represented by equirectangular projection (ERP).
Let $I^{\mathrm{ERP}} \in \mathbb{R}^{H \times W \times 3}$ denote an ERP panorama, which is obtained by sampling an underlying spherical image
$I^{\mathbb{S}} : \mathbb{S}^2 \rightarrow \mathbb{R}^3$.
We consider a user prompt specified in ERP pixel coordinates $\mathbf{p}=(u_p,v_p)\in[0,W)\times[0,H)$, and aim to predict a panoramic segmentation mask on the ERP plane
$M^{\mathrm{ERP}} \in \{0,1\}^{H \times W}$:
\begin{equation}
M^{\mathrm{ERP}} = f_{\theta}^{\mathrm{pano}}(I^{\mathrm{ERP}}, \mathbf{p}).
\end{equation}
Equivalently, $M^{\mathrm{ERP}}$ can be viewed as a spherical mask $M^{\mathbb{S}}:\mathbb{S}^2\rightarrow\{0,1\}$ through the standard ERP-to-sphere coordinate mapping.

\begin{figure}[t]
  \centering
\includegraphics[width=\textwidth]{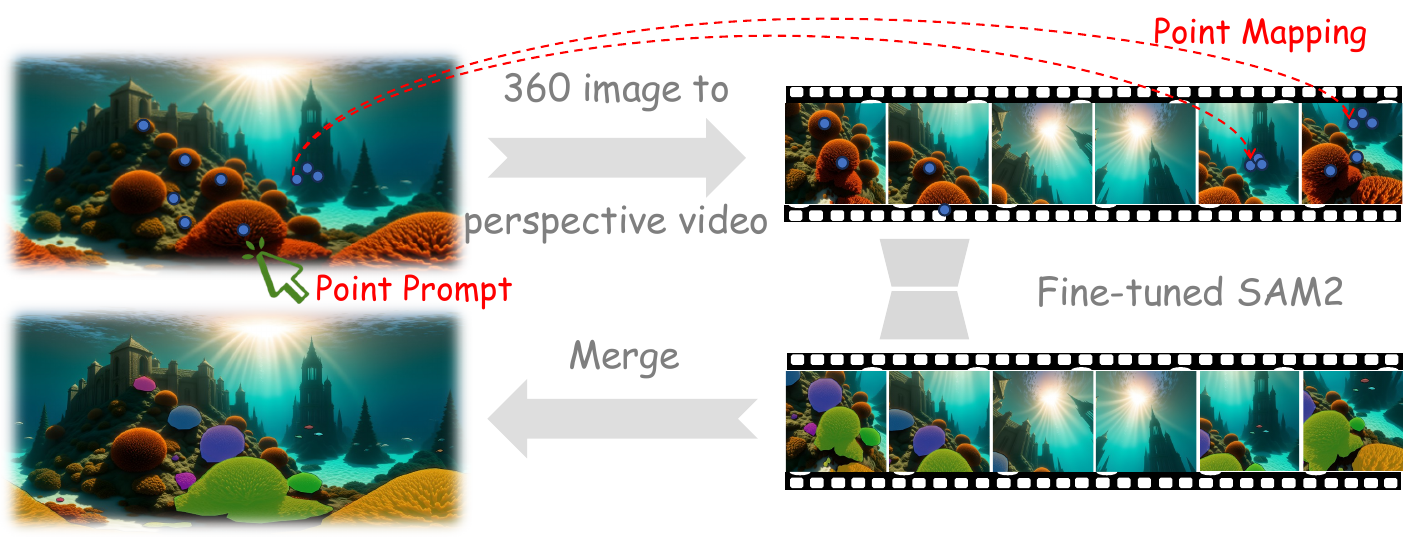}
   \caption{Overview of our pipeline. We reformulate panoramic segmentation as perspective scanning video segmentation. A panorama with the prompt points is first projected into multiple continuous perspective frames with a fixed trajectory. Next, the scanning video is processed with the fine-tuned SAM2 to generate video segmentation results, which are merged back to the ERP plane to obtain the final segmentation.}
   \label{fig:scan}
   \vspace{-15pt}
\end{figure}

\subsection{SAP Framework}

As illustrated in Fig.~\ref{fig:scan}, given user-provided prompt points, our pipeline first converts the ERP panorama into a perspective pseudo-video by following a predefined scanning trajectory. In parallel, we project the prompt points into the corresponding perspective frame using the same transformation matrix. This representation allows us to exploit SAM2’s strong prior for temporally consistent video segmentation.
However, the camera motion induced by panoramic scanning differs from the motion patterns seen in most of SAM2’s training data. To mitigate this mismatch, we fine-tune SAM2 to better adapt to the fixed-trajectory motion introduced by our scanning strategy. Finally, after obtaining per-frame masks on the generated sequence, we reproject and fuse the predicted masks back into the ERP domain to produce the final panoramic segmentation.

\subsection{Panorama to Perspective Video Frames}
Given a panorama with ERP representation $I^{\mathrm{ERP}} \in \mathbb{R}^{H \times W \times 3}$, our objective is to extract $N$ perspective image patches $I^{\mathrm{Pers}}=\{I^{\mathrm{Pers}}_1, \dots, I^{\mathrm{Pers}}_N\}$. Each patch has a spatial resolution of $L \times L$ and corresponds to a specific spherical viewpoint drawn from $\mathbf{v}=\{\mathbf{v}_1, \dots, \mathbf{v}_N\}$. The transformation from $I^{\mathrm{ERP}}$ to $I^{\mathrm{Pers}}$ is accomplished through a three-stage geometric projection.

\noindent \textbf{(i)} We first define the intrinsic and extrinsic parameters for $N$ simulated cameras. All simulated cameras share a universal intrinsic matrix $\mathbf{K}$, which relies on the focal length $f$ computed from the specified field-of-view (FoV) $\beta$ via $f = \frac{L-1}{2\tan(\beta / 2)}$. The principal point is positioned at the image center, where $c_x = c_y = \frac{L-1}{2}$. For the $i$-th simulated camera, the extrinsic matrix is denoted as \(\mathbf{E}_i = [\mathbf{R}_i, \mathbf{t}_i]\). Here, the translation vector $\mathbf{t}_i$ is assumed to be zero, while the rotation matrix $\mathbf{R}_i$ is determined by its corresponding target viewpoint $\mathbf{v}_i$.

\noindent \textbf{(ii)} We establish a 2D coordinate grid $\mathbf{X}\in\mathbb{R}^{L\times L\times 2}$ for the perspective frame, storing pixel coordinates $(u,v)$.
For each pixel, we back-project it to a ray direction in the camera coordinate system by
$\mathbf{r}^{\mathrm{cam}} \propto \mathbf{K}^{-1}[u,v,1]^T$.
We then rotate the ray to the world coordinate system using $\mathbf{R}_i$ and normalize the coordinate:
\begin{equation}
[x,y,z]^T = \operatorname{Normalize}(\mathbf{R}_i\,\mathbf{K}^{-1}[u,v,1]^T) .
\end{equation}

\noindent \textbf{(iii)} In the final stage, the derived 3D camera coordinates are converted into latitude and longitude spherical coordinates $(\theta, \phi)$ through the transformation $[\theta, \phi]^{T}=[\text{arcsin}(z),\operatorname{atan2}(y, x)]^{T}$. By utilizing these calculated angular coordinates as spatial indices, we sample the corresponding pixel values from the source panorama $I^{\mathrm{ERP}}$ to construct the final perspective views $I^{\mathrm{Pers}}$.

To cover the full sphere, we sample a discrete set of camera viewpoints and render each viewpoint as a perspective frame.
We then order these frames into a temporally coherent pseudo-video so that SAM2 can leverage its temporal prior for consistent mask propagation.
Specifically, we define $N_{\text{yaw}}$ longitudinal viewpoints and $N_{\text{pitch}}$ latitudinal viewpoints, totaling $N = N_{\text{yaw}} \times N_{\text{pitch}}$ viewpoints per panorama.
These viewpoints are organized into a temporal sequence of frames $\{\mathbf{F}_1, \dots, \mathbf{F}_N\}$.
We detail our scanning trajectory in the following subsection and describe how we cyclically shift it to support arbitrary starting viewpoints.

\noindent \textbf{Prompt point projection.}
User-provided prompt points must also be mapped onto the corresponding perspective frames.
Given a prompt point at ERP pixel coordinates $\mathbf{p} = (u_p, v_p)$, we first convert it to spherical coordinates $(\theta_p, \phi_p)$ following the standard equirectangular mapping, and then obtain its unit-sphere direction vector:
\begin{equation}
    \mathbf{d} = [\cos\theta_p\cos\phi_p,\;\cos\theta_p\sin\phi_p,\;\sin\theta_p]^T.
\end{equation}
For the $i$-th perspective frame, the direction is transformed into the camera coordinate system as $\mathbf{d}^{\mathrm{cam}} = \mathbf{R}_i^{-1}\,\mathbf{d}$, and projected onto the image plane via the shared intrinsic matrix $\mathbf{K}$ to obtain the 2D location $(\hat{u}, \hat{v})$.
The prompt is deemed visible in frame $i$ if $z_c > 0$ and $0 \leq \hat{u}, \hat{v} < L$, where $\mathbf{d}^{\mathrm{cam}} = [x_c, y_c, z_c]^T$.
By evaluating this projection across the scanning sequence, we identify all of the frames in which the prompt is visible and supply $(\hat{u}, \hat{v})$ as the input prompt to SAM2's network.

\subsection{Mask Reprojection and Fusion}
After obtaining per-frame masks $\{M_i\}_{i=1}^{N}$, we reproject them onto the ERP plane via inverse mapping.
For each panorama pixel $(u, v)$, we convert it to a unit-sphere direction $\mathbf{d}$, transform it into the $i$-th camera frame as $\mathbf{d}^{\mathrm{cam}} = \mathbf{R}_i^{-1}\mathbf{d}$, and project it onto the viewport via $\mathbf{K}$ to obtain the sampling coordinate.
The reprojected masks are fused with an element-wise maximum across all overlapping frames:
\begin{equation}
    M^{\mathrm{ERP}}(u,v) = \max_{i:\,(u,v)\in \mathcal{V}_i} \; \tilde{M}_i(u,v),
\end{equation}
where $\mathcal{V}_i$ denotes the set of panorama pixels visible in frame $i$ and $\tilde{M}_i$ is the bilinearly sampled mask value from frame $i$.

\subsection{Scanning Trajectory}

\begin{wrapfigure}{R}{0pt}
    \centering
    \includegraphics[width=0.4\linewidth]{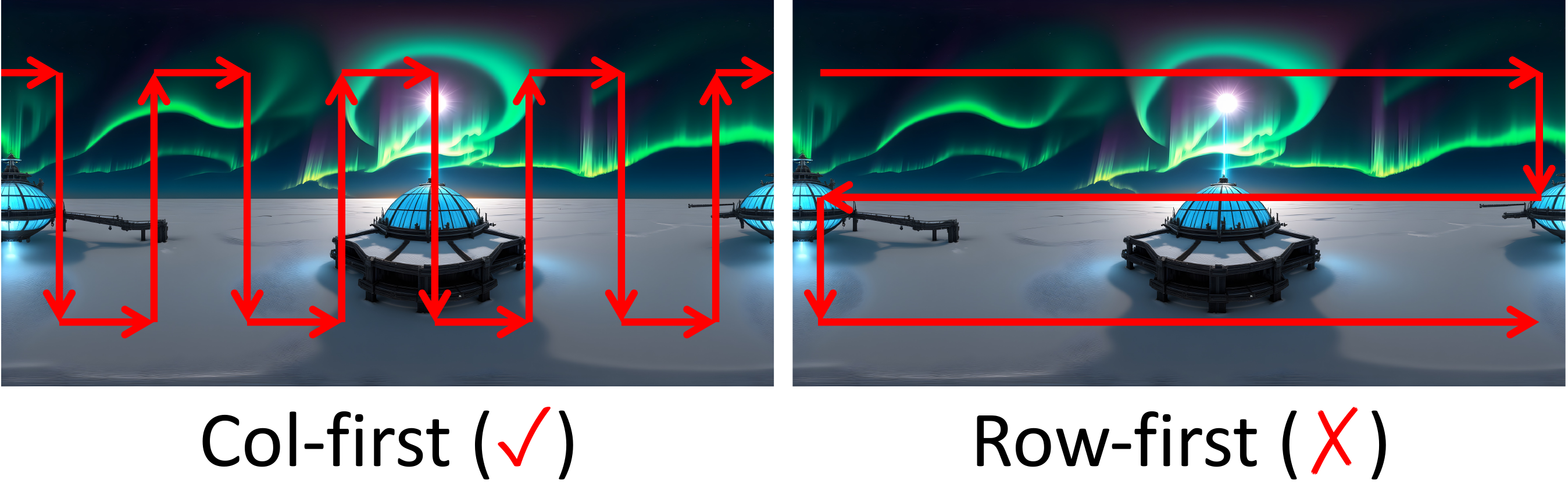}
    \vspace{-10pt}
    \caption{The visualization of different trajectories.}
    \label{fig: trajectory}
\end{wrapfigure}

Our key design requirement is that the scanning trajectory must support an infinite loop: \textbf{from any starting point}, the scan should be able to return to that point accurately while utilizing the exact same trajectory. This requirement motivates our choice of a column-first traversal, since a row-first traversal cannot satisfy it. 
As shown in Fig.~\ref{fig: trajectory}, column-first scanning can cycle indefinitely via reciprocating up-and-down motions, allowing an arbitrary starting point. In contrast, for row-first scanning, starting from an non-left-top position cannot return to the initial point after arriving the right-bottom.

\noindent \textbf{Grid construction.}
Given the horizontal FoV $\beta_h$ and vertical FoV $\beta_v$, we define the angular step between adjacent viewpoints as
\begin{equation}
    \Delta_{\text{yaw}} = \beta_h \,(1 - r), \qquad
    \Delta_{\text{pitch}} = \beta_v \,(1 - r),
\end{equation}
where $r \in (0, 1)$ is an overlap ratio controlling the spatial overlap between consecutive frames. This yields a sampling grid of
\begin{equation}
    N_{\text{yaw}} = \left\lceil \frac{360^{\circ}}{\Delta_{\text{yaw}}} \right\rceil, \qquad
    N_{\text{pitch}} = \left\lceil \frac{180^{\circ} - \beta_v}{\Delta_{\text{pitch}}} \right\rceil + 1
\end{equation}
columns and rows respectively, totaling $N = N_{\text{yaw}} \times N_{\text{pitch}}$ viewpoints that jointly cover the full panoramic sphere. The pitch values are uniformly distributed from $(90^{\circ} - \beta_v/2)$ to $(-90^{\circ} + \beta_v/2)$, and the yaw starts at $0^{\circ}$ and increment by $\Delta_{\text{yaw}}$.

\noindent \textbf{Column-first zigzag traversal.}
We organize the $N$ viewpoints into a temporally coherent pseudo-video via a \textit{column-first zigzag} scanning order. The camera first sweeps vertically along the initial yaw column, then advances one yaw step and sweeps vertically in the opposite direction, alternating with each successive column. Formally, for column $j \in \{1, \dots, N_{\text{yaw}}\}$, the row visitation order is
\begin{equation}
\mathcal{O}_j =
\begin{cases}
(j,1),(j,2),\dots,(j,N_{\text{pitch}}), & j \bmod 2 = 1,\\
(j,N_{\text{pitch}}),\dots,(j,2),(j,1), & j \bmod 2 = 0.
\end{cases}
\end{equation}
and the full trajectory is the concatenation $\mathcal{T} = \mathcal{O}_1 \,\|\, \mathcal{O}_2 \,\|\, \cdots \,\|\, \mathcal{O}_{N_{\text{yaw}}}$.
This design ensures that between any two consecutive frames, only a single angular dimension---either yaw or pitch---changes, yielding smooth, video-like transitions that align well with SAM2's temporal modeling assumptions. Furthermore, when $N_{\text{yaw}}$ is even, the last frame of the trajectory and the first frame reside at the same pitch and differ by only one yaw step, forming a \textit{closed loop} on the sphere.

\noindent \textbf{Cyclic extension for arbitrary starting points.}
To fulfill the arbitrary-start requirement, we exploit the naturally infinite property by replicating the trajectory 2 times end-to-end, yielding a cyclic sequence of $2 \cdot N$ frames.
During training, we randomly sample a starting index $s \in \{0, \dots, N - 1\}$ and extract a contiguous window of $N$ frames $\{\mathbf{F}_{s}, \dots, \mathbf{F}_{s+N-1}\}$. By construction, any such window covers all $N_{\text{yaw}} \times N_{\text{pitch}}$ viewpoints at least once, guaranteeing full panoramic coverage regardless of the entry point.

\noindent \textbf{Parameter settings.}
We set the scanning FoV to $\beta_h=\beta_v=90^\circ$ to balance traversal efficiency and image quality: a larger FoV would reduce the required viewpoints but introduces stronger projection distortion near the view boundaries, which empirically degrades downstream segmentation and temporal consistency. Under $\beta_v=90^\circ$, we fix the overlap ratio to $r=0.5$, yielding a uniform pitch step $\Delta_{\text{pitch}}=\beta_v(1-r)=45^\circ$ and a compact number of viewpoints. While larger $r$ values are feasible, they shrink the step size and rapidly increase the total number of sampled viewpoints (and thus the pseudo-video length), leading to an unacceptable increase in both training and inference cost.

%% file: sections/dataset.tex
\section{Dataset}

In this section, we introduce our proposed synthetic dataset for training, as well as the benchmarks used for zero-shot evaluation.

\subsection{Synthesis Dataset}

We spent \textbf{40,000 GPU hours} utilizing the InfiniGen engine~\cite{raistrick2024infinigen} to synthesize \textbf{183,440} 4K equirectangular panoramic images ($4096 \times 2048$ resolution and $360^\circ \times 180^\circ$ FoV), each paired with fine-grained instance segmentation masks. There are \textbf{6,409,732 instance masks} in total.
We follow the COCO-style~\cite{lin2014microsoft} object size stratification based on mask area and rescale the thresholds to match our 4K resolution setting. Specifically, we define \textit{small} objects as those with area in $[1,64^2]$, \textit{medium} as those in $(64^2,192^2]$, and \textit{large} as those with area $>192^2$.
According to this standard, our dataset comprises 37.84\% small objects, 25.70\% medium objects, and 36.47\% large objects.
Similar to SAM2, we randomly sample 200 panorama images as the validation set.

As discussed in the previous section, we aim for this dataset to improve the model’s ability to segment objects along a fixed trajectory, thereby advancing 4K-resolution panoramic image segmentation.

\subsection{Benchmark Dataset}

To evaluate the generality of our SAP, we use a zero-shot setting to compute the mIoU based on the following datasets.

\noindent \textbf{PAV-SOD}~\cite{zhang2023pav} is a task that targets instance segmentation in 360$^\circ$ panoramic videos by leveraging both visual content and audio cues to model human attention in immersive, real-world dynamic scenes. The dataset contains 67 4K-resolution videos in equirectangular projection (ERP). From these videos, we randomly sample three frames each, yielding 201 4K-resolution images for the evaluation of the visual promptable instance segmentation task. In total, this benchmark comprises 427 instances, including 42 small, 184 medium, and 201 large instances. This benchmark can serve as the real-world 4K panorama instance segmentation task.

\noindent \textbf{HunyuanWorld 1.0}~\cite{team2025hunyuanworld} is an open-source generative framework for creating immersive, explorable, and interactive 360$^\circ$ panorama images from text or a single reference image. In our setup, we used HunyuanWorld-1.0 to generate 100 8K-resolution panoramic images, and annotated object instance segmentation masks for randomly selected objects to evaluate how well our model generalizes to synthetic images. Overall, this benchmark comprises 264 instances, including 48 small, 113 medium, and 103 large instances. This benchmark serves as the 8K panorama segmentation task for robustness evaluation of synthesized images.

%% file: sections/exp.tex
\section{Experiments}

\subsection{Implementation Details}

We build our method on top of SAM2~\cite{ravi2024sam} with the Hiera-Large image encoder~\cite{ryali2023hiera}. Since our goal is to enhance SAM’s performance for trajectory-specific segmentation, we freeze the image encoder and update only the memory attention, memory encoder, mask decoder, and prompt encoder, thereby keeping the feature extraction capacity stable. Moreover, as our dataset contains only simulated data, we additionally mix the original training data (SA-1B and SA-V) into the training set to further mitigate catastrophic forgetting~\cite{kirkpatrick2017overcoming}.
For panoramic data, we convert each equirectangular image ($4096\times2048$) into a pseudo-video by rendering $1024\times1024$ perspective viewports with a $90^\circ$ field-of-view and 50\% overlap, traversed in a column-first snake order for 2 cycles.
We use AdamW~\cite{loshchilovdecoupled} with a batch size of 128, learning rate of $2\times10^{-4}$ (cosine schedule~\cite{loshchilov2017sgdr}), weight decay of 0.1, and gradient clipping at 0.1.

\subsection{Metrics}

We adopt mean Intersection over Union (mIoU) as our primary evaluation metric under an interactive segmentation protocol.
For each ground-truth instance, we simulate user clicks by selecting the point farthest from the mask boundary as the initial positive prompt.
The per-instance IoU is computed as $\text{IoU} = |P \cap G| / |P \cup G|$, where $P$ and $G$ denote the predicted and ground-truth binary masks, respectively.
As the protocol~\cite{sofiiuk2022reviving, sofiiuk2020f} of the SAM series is applied, we report results under two settings:

\begin{itemize}
    \item \textbf{1-click mIoU}: Each instance receives a single positive click at the point farthest from its boundary. The model produces a segmentation from this single prompt, and mIoU is computed by averaging IoU over all instances.
    \item \textbf{3-click Error Correction mIoU}: Starting from the 1-click prediction, we perform iterative error correction. At each subsequent round, we compare the current prediction with the ground truth to identify the largest error region, including false negative (FN) and false positive (FP). A corrective click is placed at the point farthest from the boundary within the larger error region, with a positive label for FN regions and a negative label for FP regions. The model then re-segments using all accumulated prompts.
\end{itemize}

\subsection{Quantitative Comparison}

\begin{table}[t!]
\centering
\tabcolsep=0.1cm
\caption{\textbf{Zero-shot} Segmentation Mask mIoU on PAV-SOD~\cite{zhang2023pav} benchmark.}
\vspace{-10pt}
\setlength{\fboxsep}{2pt}
\resizebox{\linewidth}{!}{
\begin{tabular}{l|cccc|cccc}
\toprule
\rowcolor{gray!15}
                & \multicolumn{4}{c|}{1-click}     & \multicolumn{4}{c}{3-click Error Correction} \\ \hline
\rowcolor{gray!15}
Method          & Overall & Small & Medium & Large & Overall    & Small    & Medium    & Large    \\ \hline
SAM2-tiny       & 51.6    & 46.3  & 55.5   & 49.1  & 82.2       & 61.7     & 79.6      & 88.8     \\
SAM2-tiny+scan  & 65.1    & 49.6  & 63.4   & 70.0  & 83.0       & 65.8     & 82.3      & 87.2     \\
SAP-tiny        & 75.8    & 53.9  & 76.6   & 79.7  & 84.8       & 68.7     & 84.7      & 88.4     \\
\rowcolor{yellow!15}
$\Delta$ (SAP-SAM2) & +24.2 & +7.6 & +21.1 & +30.6 & +2.6 & +7.0 & +5.1 & -0.4 \\ \hline
SAM2-small      & 54.0    & 45.8  & 59.7   & 50.6  & 83.1       & 60.8     & 80.3      & 90.3     \\
SAM2-small+scan & 67.4    & 53.7  & 63.0   & 74.3  & 84.0       & 69.4     & 81.7      & 89.1     \\
SAP-small       & 72.0    & 51.7  & 69.7   & 78.3  & 84.3       & 67.8     & 83.2      & 88.8     \\
\rowcolor{yellow!15}
$\Delta$ (SAP-SAM2) & +18.0 & +5.9 & +10.0 & +27.7 & +1.2 & +7.0 & +2.9 & -1.5 \\ \hline
SAM2-base       & 60.4    & 49.9  & 67.5   & 56.0  & 84.6       & 65.2     & 82.2      & 90.8     \\
SAM2-base+scan  & 67.9    & 54.8  & 64.1   & 74.0  & 83.5       & 68.2     & 81.8      & 88.2     \\
SAP-base        & 76.1    & 56.9  & 75.1   & 81.1  & 84.2       & 70.0     & 81.9      & 89.2     \\
\rowcolor{yellow!15}
$\Delta$ (SAP-SAM2) & +15.7 & +7.0 & +7.6 & +25.1 & -0.4 & +4.8 & -0.3 & -1.6 \\ \hline
SAM2-large      & 66.3    & 50.7  & 71.9   & 64.4  & 84.3       & 65.5     & 81.7      & 90.5     \\
SAM2-large+scan & 69.0    & 58.4  & 66.2   & 73.8  & 84.1       & 68.4     & 81.6      & 89.5     \\
SAP-large       & 77.3    & 61.1  & 76.2   & 81.7  & 86.1       & 71.8     & 84.3      & 90.6    \\ 
\rowcolor{yellow!15}
$\Delta$ (SAP-SAM2) & +11.0 & +10.4 & +4.3 & +17.3 & +1.8 & +6.3 & +2.6 & +0.1 \\ \bottomrule
\end{tabular}}
\vspace{-10pt}
\label{tab: PAV-SOD}
\end{table}

\begin{table}[t!]
\centering
\tabcolsep=0.1cm
\caption{\textbf{Zero-shot} Segmentation Mask mIoU on HunyuanWorld-1.0~\cite{team2025hunyuanworld} benchmark.}
\vspace{-10pt}
\setlength{\fboxsep}{2pt}
\resizebox{\linewidth}{!}{
\begin{tabular}{l|cccc|cccc}
\toprule
\rowcolor{gray!15}
                & \multicolumn{4}{c|}{1-click}     & \multicolumn{4}{c}{3-click Error Correction} \\ \hline
\rowcolor{gray!15}
Method          & Overall & Small & Medium & Large & Overall    & Small    & Medium    & Large    \\ \hline
SAM2-tiny       & 68.6    & 77.6  & 76.7   & 56.0  & 84.7       & 82.9     & 85.9      & 84.4     \\
SAM2-tiny+scan  & 46.7    & 40.0  & 52.9   & 43.1  & 78.7       & 75.5     & 81.0      & 77.7     \\
SAP-tiny        & 75.0    & 73.3  & 80.8   & 69.6  & 85.7       & 83.5     & 85.2      & 87.3     \\
\rowcolor{yellow!15}
$\Delta$ (SAP-SAM2) & +6.4 & -4.3 & +4.1 & +13.6 & +1.0 & +0.6 & -0.7 & +2.9 \\ \hline
SAM2-small      & 68.7    & 77.7  & 76.7   & 56.1  & 86.3       & 82.6     & 86.2      & 87.9     \\
SAM2-small+scan & 45.6    & 40.2  & 50.0   & 43.3  & 80.6       & 76.2     & 84.0      & 79.0     \\
SAP-small       & 74.3    & 72.9  & 78.6   & 70.2  & 85.9       & 83.1     & 87.2      & 85.8     \\
\rowcolor{yellow!15}
$\Delta$ (SAP-SAM2) & +5.6 & -4.8 & +1.9 & +14.1 & -0.4 & +0.5 & +1.0 & -2.1 \\ \hline
SAM2-base       & 70.6    & 78.5  & 79.2   & 58.0  & 85.7       & 82.8     & 87.6      & 84.9     \\
SAM2-base+scan  & 45.5    & 39.5  & 47.7   & 46.0  & 80.5       & 74.5     & 83.8      & 79.7     \\
SAP-base        & 77.4    & 78.0  & 82.3   & 71.7  & 85.8       & 83.8     & 87.9      & 84.5     \\
\rowcolor{yellow!15}
$\Delta$ (SAP-SAM2) & +6.8 & -0.5 & +3.1 & +13.7 & +0.1 & +1.0 & +0.3 & -0.4 \\ \hline
SAM2-large      & 70.8    & 78.8  & 78.6   & 58.8  & 82.1       & 83.3     & 83.7      & 79.9     \\
SAM2-large+scan & 54.9    & 49.4  & 58.8   & 53.2  & 81.5       & 74.7     & 83.6      & 82.3     \\
SAP-large       & 77.4    & 77.1  & 82.9   & 71.5  & 88.3       & 84.0     & 89.3      & 89.3     \\
\rowcolor{yellow!15}
$\Delta$ (SAP-SAM2) & +6.6 & -1.7 & +4.3 & +12.7 & +6.2 & +0.7 & +5.6 & +9.4 \\ \bottomrule
\end{tabular}}
\vspace{-10pt}
\label{tab: HunyuanWorld}
\end{table}

\begin{table}[t!]
\centering
\tabcolsep=0.1cm
\caption{Segmentation mask mIoU on InfiniGen~\cite{raistrick2024infinigen} benchmark.}
\vspace{-10pt}
\setlength{\fboxsep}{2pt}
\resizebox{\linewidth}{!}{
\begin{tabular}{l|cccc|cccc}
\toprule
\rowcolor{gray!15}
                & \multicolumn{4}{c|}{1-click}     & \multicolumn{4}{c}{3-click Error Correction} \\ \hline
\rowcolor{gray!15}
Method          & Overall & Small & Medium & Large & Overall    & Small    & Medium    & Large    \\ \hline
SAM2-tiny       & 59.7    & 55.3  & 69.0   & 57.5  & 80.2       & 67.7     & 84.4      & 90.0     \\
SAM2-tiny+scan  & 59.4    & 60.4  & 60.8   & 57.0  & 83.8       & 78.4     & 86.6      & 87.7     \\
SAP-tiny        & 77.3    & 64.9  & 82.1   & 87.6  & 87.4       & 79.4     & 90.7      & 93.8     \\
\rowcolor{yellow!15}
$\Delta$ (SAP-SAM2) & +17.6 & +9.6 & +13.1 & +30.1 & +7.2 & +11.7 & +6.3 & +3.8 \\ \hline
SAM2-small      & 59.8    & 57.2  & 68.0   & 56.6  & 80.5       & 67.5     & 84.7      & 91.0     \\
SAM2-small+scan & 61.4    & 62.0  & 62.4   & 60.0  & 84.5       & 78.4     & 86.7      & 89.7     \\
SAP-small       & 79.0    & 68.5  & 82.8   & 88.0  & 87.5       & 79.7     & 90.9      & 93.8     \\
\rowcolor{yellow!15}
$\Delta$ (SAP-SAM2) & +19.2 & +11.3 & +14.8 & +31.4 & +7.0 & +12.2 & +6.2 & +2.8 \\ \hline
SAM2-base       & 62.0    & 57.6  & 71.2   & 59.8  & 81.4       & 68.2     & 86.0      & 91.6     \\
SAM2-base+scan  & 56.6    & 55.9  & 56.5   & 57.3  & 83.5       & 77.1     & 85.8      & 88.8     \\
SAP-base        & 81.8    & 72.3  & 85.8   & 89.6  & 88.9       & 82.3     & 91.8      & 94.1     \\
\rowcolor{yellow!15}
$\Delta$ (SAP-SAM2) & +19.8 & +14.7 & +14.6 & +29.8 & +7.5 & +14.1 & +5.8 & +2.5 \\ \hline
SAM2-large      & 62.8    & 59.7  & 69.8   & 60.7  & 81.4       & 69.7     & 85.4      & 90.6     \\
SAM2-large+scan & 65.4    & 63.9  & 66.5   & 66.2  & 84.2       & 77.9     & 86.3      & 89.8     \\
SAP-large       & 81.9    & 72.5  & 84.5   & 90.7  & 89.0       & 81.5     & 92.2      & 95.1     \\
\rowcolor{yellow!15}
$\Delta$ (SAP-SAM2) & +19.1 & +12.8 & +14.7 & +30.0 & +7.6 & +11.8 & +6.8 & +4.5 \\ \bottomrule
\end{tabular}}
\vspace{-10pt}
\label{tab: InfiniGen}
\end{table}

To enable a quantitative comparison with the baselines, we follow the SAM series~\cite{kirillov2023segment, ravi2024sam} and evaluate all models using the same metrics: 1-click mIoU and 3-click error-correction mIoU.
We adopt SAM2 as the primary baseline and report results using the official checkpoints of four model sizes: Tiny, Small, Base, and Large.
In addition, we include SAM2 equipped with our proposed “scan” strategy (training-free) as an additional baseline.
Finally, we report the results of our full SAP method for four model sizes.
As shown in Tab.~\ref{tab: PAV-SOD} - \ref{tab: InfiniGen}, our approach yields consistent and substantial improvements, with notable gains on the 4K/8K benchmarks. 
Please note that the PAV-SOD and HunyuanWorld-1.0 benchmarks are \textbf{zero-shot} settings.

\noindent \textbf{PAV-SOD Benchmark.}
As demonstrated in Tab.~\ref{tab: PAV-SOD}, we evaluate our SAP and baselines on real-world 4K panoramic images in a zero-shot setting. 
With our perspective scanning strategy, SAM2 already benefits from improved cross-view continuity, yielding noticeable gains, particularly for smaller models (e.g., +13.5 mIoU on the tiny variant). 
After fine-tuning SAM2 on our simulated data using the same scanning trajectory, SAP achieves larger and more consistent improvements across all model scales (an average mIoU gain of +17.2 across the four model sizes), suggesting that fixed-trajectory training is crucial for stable temporal memory usage under the fixed scanning motion. Especially for the tiny model, we achieve +24.2 mIoU improvement.

\begin{figure}[t!]
    \centering
    \includegraphics[width=1\linewidth]{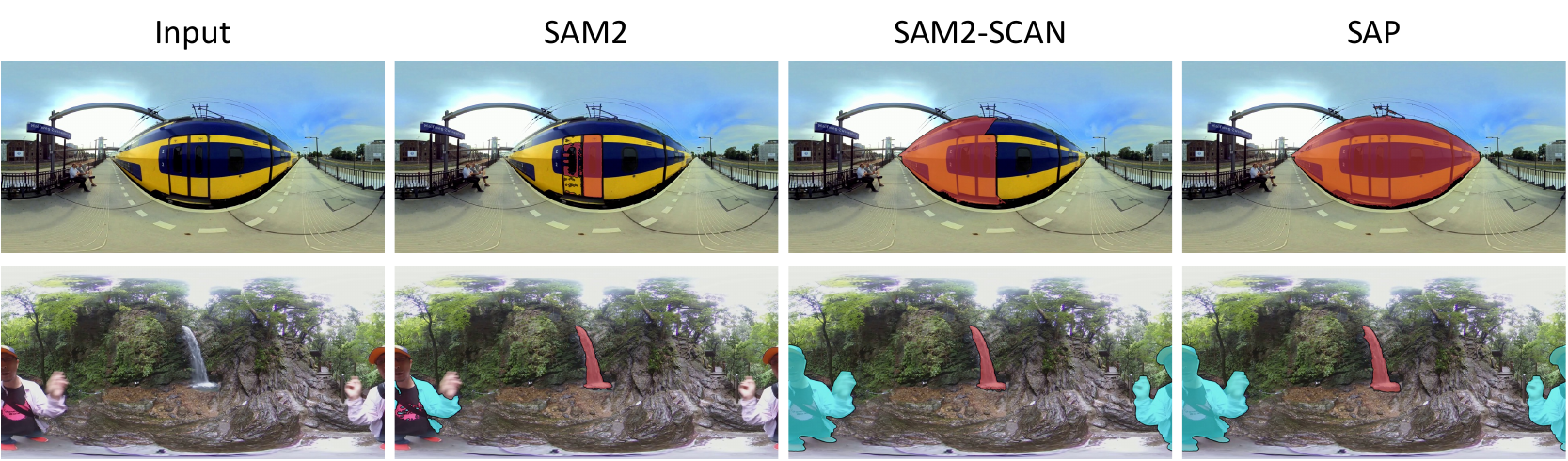}
    \vspace{-10pt}
    \caption{Qualitative Comparison based on PAV-SOD~\cite{zhang2023pav} dataset.}
    \vspace{-10pt}
    \label{fig: qualitative PAV-SOD}
\end{figure}
\begin{figure}[t!]
    \centering
    \includegraphics[width=1\linewidth]{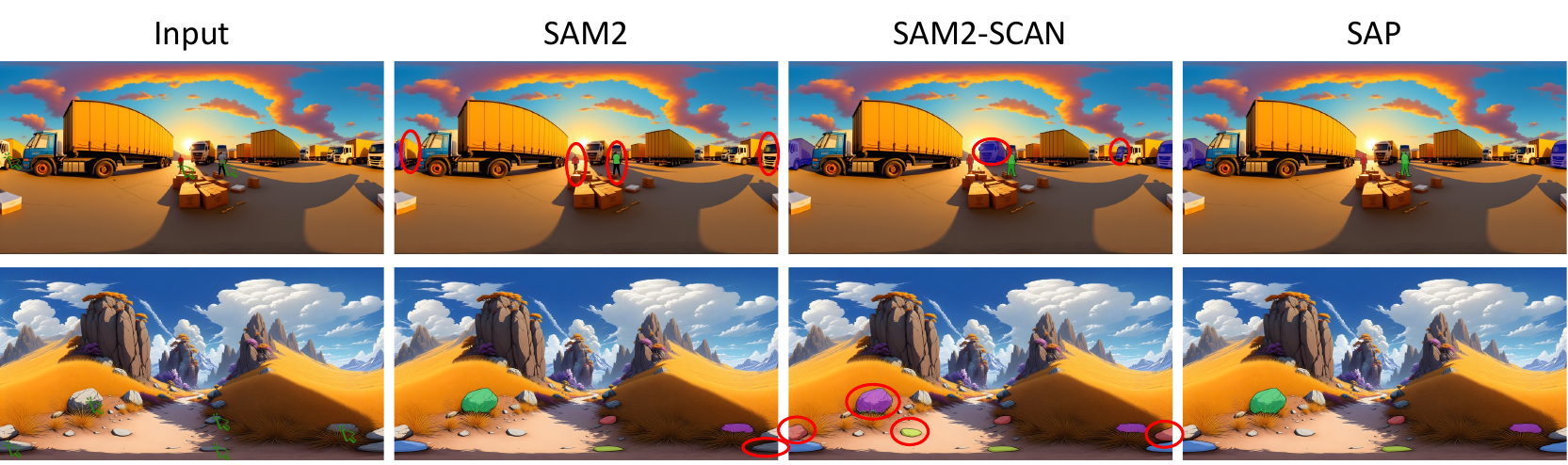}
    \vspace{-10pt}
    \caption{Qualitative Comparison based on HunyuanWorld-1.0~\cite{team2025hunyuanworld} dataset. The red circles indicate segmentation errors.}
    \vspace{-10pt}
    \label{fig: qualitative HY}
\end{figure}

\noindent \textbf{HunyuanWorld-1.0 Benchmark.}
As shown in Tab.~\ref{tab: HunyuanWorld}, to assess the robustness of SAP, we introduce the HunyuanWorld-1.0 benchmark with manually annotated instance masks.
The HunyuanWorld panoramas exhibit a strong cartoon-style appearance shift that is largely absent from SAM2’s video training data. As a result, directly applying the scanning strategy to SAM2 without trajectory-aligned training leads to a substantial performance drop. In contrast, after fine-tuning with the fixed scanning trajectory, SAP consistently improves over SAM2 across model scales, indicating that fine-tuning is essential for making the temporal memory mechanism robust to synthetic domain shifts and for maintaining cross-view mask consistency.

\noindent \textbf{InfiniGen Benchmark.}
As shown in Tab.~\ref{tab: InfiniGen}, we report mIoU on our InfiniGen 4K panoramic benchmark.
Overall, SAP outperforms both vanilla SAM2 and the scanning baseline across model scales.
Though enhancing SAM2 with our training-free scanning strategy can promote the performance slightly, SAP can use the streaming memory to propagate instance identity and mask boundaries across adjacent views, which is important for large spatial extent and cross-view continuity in 4K panoramas by fine-tuning SAM2 under the same trajectory. We observe the same trend in the real-world zero-shot setting.

\subsection{Qualitative Comparison}

As illustrated in Fig.~\ref{fig: qualitative PAV-SOD} and Fig.~\ref{fig: qualitative HY}, we present qualitative comparisons on two zero-shot benchmarks, PAV-SOD~\cite{zhang2023pav} and HunyuanWorld-1.0~\cite{team2025hunyuanworld}. In Fig.~\ref{fig: qualitative PAV-SOD}, vanilla SAM2 is severely affected by large-object distortions (first row) and struggles to segment the seam region between the two halves of a panorama (second row). As shown in Fig.~\ref{fig: qualitative HY}, vanilla SAM2 also fails around the seam area, while SAM2-SCAN tends to produce excessive masks for visually similar objects. After fine-tuning, our SAP delivers high-quality and accurate segmentation results on 4K/8K panoramic images.

\subsection{Ablation Studies}

In this section, we discuss our design choices in five groups of ablation studies. All of these experiments are based on the ``large'' model and 1-click mIoU.

\begin{table}[t!]
\centering
\tabcolsep=0.2cm
\caption{The ablation studies on the training sets.}
\vspace{-10pt}
\resizebox{\linewidth}{!}{
\begin{tabular}{cc|cccc|cccc}
\toprule
\multicolumn{2}{c|}{Training Data} & \multicolumn{4}{c|}{PAV-SOD}     & \multicolumn{4}{c}{HunyuanWorld-1.0}    \\ \hline
SAM2           & Ours          & Overall & Small & Medium & Large & Overall   & Small  & Medium  & Large   \\ \hline
$\checkmark$   &               & 60.0    & 58.9  & 60.6   & 59.7  & 47.4      & 46.8   & 51.5    & 43.3     \\
               & $\checkmark$  & 67.3    & 44.2  & 59.4   & 79.2  & 71.8      & 63.8   & 77.8    & 69.1     \\
$\checkmark$   & $\checkmark$  & 77.3    & 61.1  & 76.2   & 81.7  & 77.4      & 77.1   & 82.9    & 71.5    \\ \bottomrule
\end{tabular}}
\vspace{-5pt}
\label{tab: ablation training set}
\end{table}

\begin{table}[t!]
\centering
\tabcolsep=0.2cm
\caption{The ablation studies of the training strategies.}
\vspace{-10pt}
\resizebox{\linewidth}{!}{
\begin{tabular}{cc|cccc|cccc}
\toprule
\multicolumn{2}{c|}{Strategy} & \multicolumn{4}{c|}{PAV-SOD}     & \multicolumn{4}{c}{HunyuanWorld-1.0}    \\ \hline
Scan          & Overlap       & Overall & Small & Medium & Large & Overall   & Small  & Medium  & Large   \\ \hline
              &               & 73.0    & 50.0  & 72.5   & 78.3  & 76.1      & 79.8   & 80.2    & 70.2     \\
$\checkmark$  &               & 66.5    & 59.4  & 65.6   & 68.6  & 72.0      & 73.2   & 79.8    & 62.5     \\
$\checkmark$  & $\checkmark$  & 77.3    & 61.1  & 76.2   & 81.7  & 77.4      & 77.1   & 82.9    & 71.5    \\ \bottomrule
\end{tabular}}
\vspace{-5pt}
\label{tab: ablation strategy}
\end{table}

\begin{table}[t!]
\centering
\tabcolsep=0.15cm
\caption{The ablation studies on the perspective transformation.}
\vspace{-10pt}
\resizebox{\linewidth}{!}{
\begin{tabular}{c|cccc|cccc}
\toprule
\multirow{2}{*}{Method} & \multicolumn{4}{c|}{PAV-SOD}     & \multicolumn{4}{c}{HunyuanWorld-1.0} \\ \cline{2-9} 
                        & Overall & Small & Medium & Large & Overall  & Small  & Medium  & Large  \\ \hline
Direct Patchify         & 72.6    & 61.2  & 72.6   & 74.9  & 76.4     & 79.7   & 82.8    & 68.0   \\
Perspective Transform   & 77.3    & 61.1  & 76.2   & 81.7  & 77.4     & 77.1   & 82.9    & 71.5    \\ \bottomrule
\end{tabular}}
\vspace{-5pt}
\label{tab: ablation perspective}
\end{table}

\noindent \textbf{Training Dataset.}
As shown in Tab.~\ref{tab: ablation training set}, we validate the effectiveness of our dataset and training strategy through ablations using (i) \textit{only SAM2 data}, (ii) \textit{only our data}, and (iii) \textit{SAM2 + our data}. The results show that incorporating our synthetic panoramas substantially improves performance on panoramic segmentation. Meanwhile, training solely on the synthetic data is insufficient. By mixing the original SAM2 training set with our synthetic data, we alleviate catastrophic forgetting~\cite{kirkpatrick2017overcoming} and maintain stable zero-shot performance.

\noindent \textbf{Finetuning without Scanning.} To further validate the core idea of perspective-scanning videos, we fine-tune SAM2 on a mixture of our collected original panoramas and the SAM2 training set. During training, we treat each panorama as a regular image by directly resizing it to 1K resolution via interpolation. As shown in Tab.~\ref{tab: ablation strategy} (Rows 1 and 3), our overlapped perspective-scanning strategy yields a substantial improvement in mIoU, demonstrating the effectiveness of SAP for ultra-high-resolution panoramic segmentation.

\noindent \textbf{Overlap of Scanning Video.}
When constructing a fixed-trajectory perspective video from an ERP panorama, we introduce overlap between adjacent perspective frames to ensure smoother transitions during spherical scanning. As shown in Tab.~\ref{tab: ablation strategy} (Rows 2 and 3), adding overlap substantially improves mIoU, highlighting its benefit for stable training. Moreover, these results suggest that without inter-frame overlap, the SAM2 memory bank cannot operate effectively.

\noindent \textbf{Perspective Transformation.} OmniSAM~\cite{zhong2025omnisam} proposes a straightforward approach that patchifies an ERP panorama—without any perspective transformation—using a sliding window for semantic segmentation. We adapt this idea for an ablation comparison. Specifically, we also perform cyclic scanning, but operate directly in the ERP domain. As shown in Tab.~\ref{tab: ablation perspective}, our perspective-scanning video achieves higher mIoU than direct ERP patchification on real-world 360$^\circ$ images, highlighting the importance of reducing the domain gap between ERP panoramas and perspective images.

\begin{wraptable}{R}{0.5\linewidth}
    \centering
    \tabcolsep=0.3cm
    \resizebox{\linewidth}{!}{
    \begin{tabular}{c|ccc}
    \toprule
    Method & SAM2 & SAM2+scan & SAP  \\ \hline
    mIoU   & 46.1 & 53.8      & 68.2    \\ \bottomrule
    \end{tabular}}
    \vspace{-5pt}
    \caption{Seam/Pole subset evaluation.}
    \vspace{5pt}
    \label{tab: seam pole}
\end{wraptable}

\noindent \textbf{Seam/Pole Area Results.} To validate our claim that SAM2~\cite{ravi2024sam} struggles with the seam/pole area, we extracted such instances (in total of 76 objects) in PAV-SOD and HunyuanWorld-1.0 zero-shot benchmarks as a subset for evaluation. As shown in Tab.~\ref{tab: seam pole}, the performance of vanilla SAM2 drops significantly in this subset of the benchmarks. Benefiting from the scanning strategy, we make the seam area connected and make the pole areas become a normal shape. Therefore, we achieve great improvement in such extreme examples.

%% file: sections/conclusion.tex
\section{Conclusion}

We introduced \textbf{Segment Any 4K Panorama (SAP)} for promptable panoramic instance segmentation by reformulating panorama segmentation as fixed-trajectory perspective video segmentation and fine-tuning SAM2 with our collected 183K large-scale synthetic instance-level dataset from the InfiniGen engine.
Our evaluation covers both in-domain benchmarking on InfiniGen and a fully \textbf{zero-shot} test suite spanning diverse panorama regimes: \textbf{real-world} 4K ultra-HD panoramas (PAV-SOD) and 8K cartoon-style panoramas generated by a text-to-panorama model (HunyuanWorld-1.0).
Across all these settings, SAP consistently improves performance, demonstrating strong generalization of our SAP to varied domains and styles.

\noindent \textbf{Future work.}
A natural next step is to extend SAP from single panoramas to panoramic videos, leveraging temporal coherence across time in addition to our intra-panorama scanning trajectory. This may enable more stable long-term instance tracking and segmentation under real camera motion.

%% file: sections/appendix.tex
\appendix

\section{Inference Pseudo Code}

For clarity and reproducibility, Algorithm~\ref{alg:panorama_interactive_seg} summarizes the 
complete inference pipeline of our method. Given an ERP panorama $I^{\mathrm{ERP}}$ and a user 
prompt point $\mathbf{p}$, the algorithm proceeds in five stages.

\begin{algorithm}[h]
\caption{Inference Pipeline}
\label{alg:panorama_interactive_seg}
\KwIn{ERP panorama $I^{\mathrm{ERP}}$, prompt point $\mathbf{p}=(u_p, v_p)$ on the ERP plane}
\KwOut{Panoramic segmentation mask $M^{\mathrm{ERP}}$}

\tcp{Step 1: Pre-cut $N$ perspective frames along scanning trajectory}
$\{\mathbf{v}_i\}_{i=0}^{N-1} \leftarrow \textsc{GenerateTrajectory}(N)$\;
\ForEach{$i \in \{0, \dots, N-1\}$}{
    $\mathbf{F}_i \leftarrow \textsc{PanoToPersp}(I^{\mathrm{ERP}}, \mathbf{v}_i)$\;
}

\tcp{Step 2: Project prompt point to all visible frames}
$\mathcal{K} \leftarrow \emptyset$\;
\ForEach{$i \in \{0, \dots, N-1\}$}{
    $(\hat{u}_i, \hat{v}_i) \leftarrow \textsc{PromptProject}(\mathbf{p}, \mathbf{v}_i)$\;
    \If{$(\hat{u}_i, \hat{v}_i)$ is within viewport}{
        $\mathcal{K} \leftarrow \mathcal{K} \cup \{(i, \hat{u}_i, \hat{v}_i)\}$\;
    }
}

\tcp{Step 3: Choose start frame and reorder to $N$-frame video}
$k' \leftarrow \min_{(i,\cdot,\cdot) \in \mathcal{K}} i$\;
$\textit{video} \leftarrow [\mathbf{F}_{k'}, \mathbf{F}_{k'+1}, \dots, \mathbf{F}_{N-1}, \mathbf{F}_0, \dots, \mathbf{F}_{k'-1}]$\;

\tcp{Step 4: Run SAM2 video segmentation}
$\textit{state} \leftarrow \textsc{SAM2.Init}(\textit{video})$\;
\ForEach{$(i, \hat{u}_i, \hat{v}_i) \in \mathcal{K}$}{
    $\textsc{SAM2.AddPoint}(\textit{state}, \textit{frame}\!=\!(i - k') \bmod N, \textit{point}\!=\!(\hat{u}_i, \hat{v}_i))$\;
}
$\{M_j\}_{j=0}^{N-1} \leftarrow \textsc{SAM2.Propagate}(\textit{state})$\;

\tcp{Step 5: Fuse perspective masks back to panorama}
$M^{\mathrm{ERP}} \leftarrow \mathbf{0}$\;
\ForEach{$j \in \{0, \dots, N-1\}$}{
    $M^{\mathrm{ERP}} \leftarrow \max\!\big(M^{\mathrm{ERP}},\ \textsc{PerspToERP}(M_j)\big)$\;
}
\Return{$M^{\mathrm{ERP}}$}\;
\end{algorithm}

\section{Analysis about the Number of Frames}

In Sec.~3.5, we introduce the formal representation of the proposed scanning trajectory. In this appendix, we further discuss practical parameter choices for constructing the pseudo-video. Specifically, we assume $\beta_h = \beta_w = 90^\circ$ and set the angular step sizes in the horizontal and vertical directions to be identical, i.e., $\Delta = \Delta_{\text{yaw}} = \Delta_{\text{pitch}}$. Under this setting, Tab.~\ref{tab:delta_overlap} reports the corresponding numbers of sampled views in the pitch and yaw directions, the total number of frames in the pseudo-video, and the overlap ratio between adjacent frames for different values of $\Delta$.

As the angular interval $\Delta$ decreases, the scanning trajectory becomes denser, resulting in a larger number of sampled views and a higher overlap ratio between neighboring frames. This denser sampling is beneficial for improving spatial continuity across frames and providing richer inter-frame correspondence. However, it also leads to a rapid increase in the total number of frames $N$. For example, when $\Delta$ decreases from $90^\circ$ to $45^\circ$, the total number of views increases from 8 to 24, while further reducing $\Delta$ to $30^\circ$ doubles the number of views again to 48. Therefore, although smaller step sizes can potentially provide more redundant contextual information, they also impose substantially higher computational and memory costs.

\begin{table}[h]
\centering
\tabcolsep=0.3cm
\caption{The feasible parameters for the trajectory of pseudo-video.}
\resizebox{0.6\linewidth}{!}{
\begin{tabular}{l|cccc}
\toprule
$\Delta$ & $N_{\text{pitch}}$ & $N_{\text{yaw}}$ & $N$ & Overlap Ratio \\
\hline
90$^\circ$   & 2 & 4  & 8   & 0.0 \\
45$^\circ$   & 3 & 8  & 24  & 0.5 \\
30$^\circ$   & 4 & 12 & 48  & 2/3 \\
22.5$^\circ$ & 5 & 16 & 80  & 0.75 \\
18$^\circ$   & 6 & 20 & 120 & 0.8 \\
\bottomrule
\end{tabular}}
\label{tab:delta_overlap}
\end{table}

In practice, the choice of $\Delta$ is constrained by available GPU resources. In our current implementation, we can only support the first two configurations in Tab.~\ref{tab:delta_overlap}, namely $\Delta=90^\circ$ and $\Delta=45^\circ$. These two settings provide a reasonable trade-off between frame overlap and computational efficiency. In particular, $\Delta=90^\circ$ corresponds to non-overlapping views and yields the minimum number of frames, while $\Delta=45^\circ$ introduces 50\% overlap between adjacent frames, which improves continuity across the pseudo-video while remaining computationally affordable.

\section{Dataset Details}

\begin{figure}[h]
    \centering
    \includegraphics[width=1\linewidth]{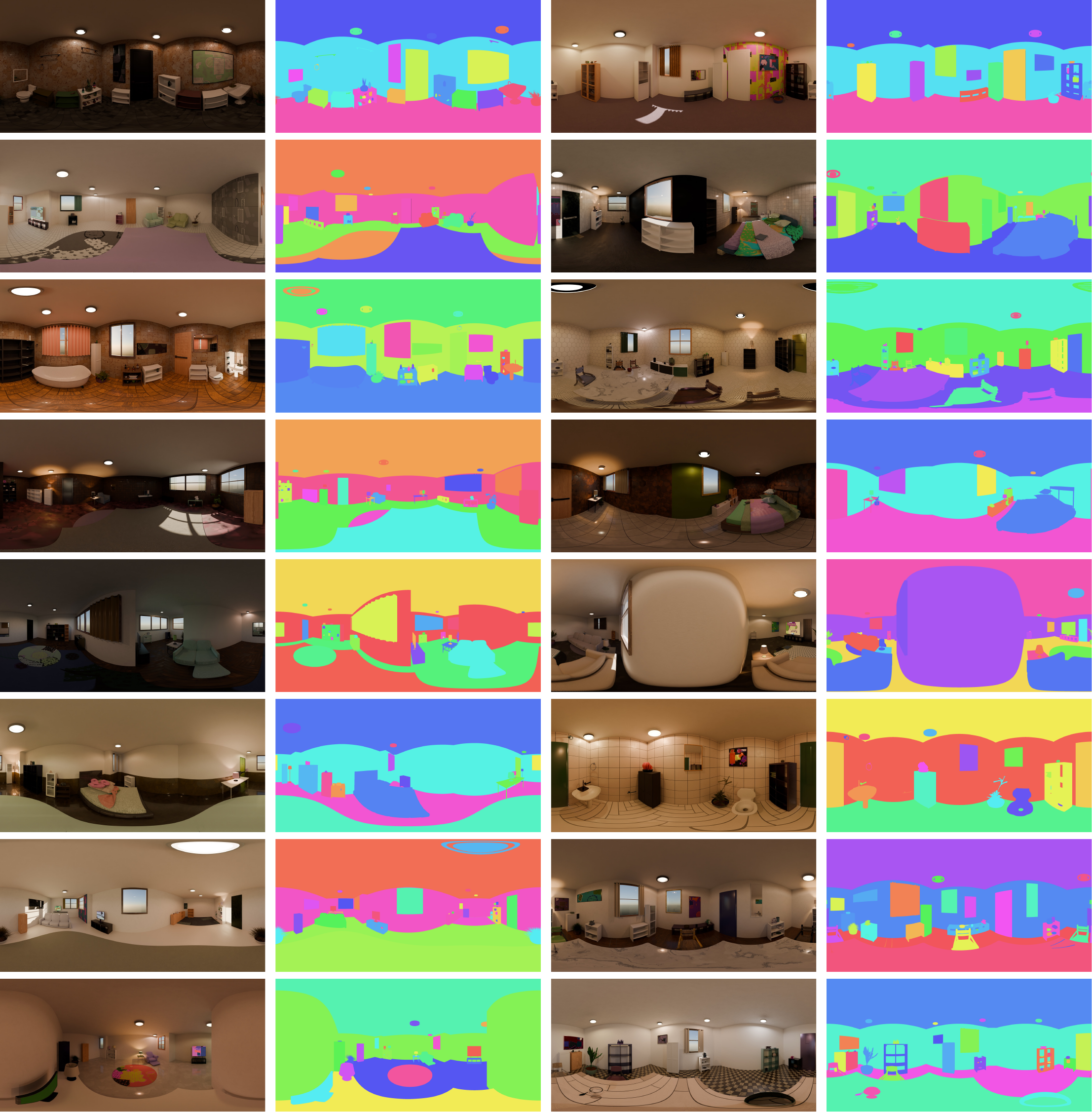}
    \caption{Demos of our collected dataset.}
    \label{fig:supple_dataset}
\end{figure}

As shown in Fig.~\ref{fig:supple_dataset}, we provide more visualization demos of our collected dataset using InfiniGen.

\section{More Qualitative Results}

\begin{figure}[h]
    \centering
    \includegraphics[width=1\linewidth]{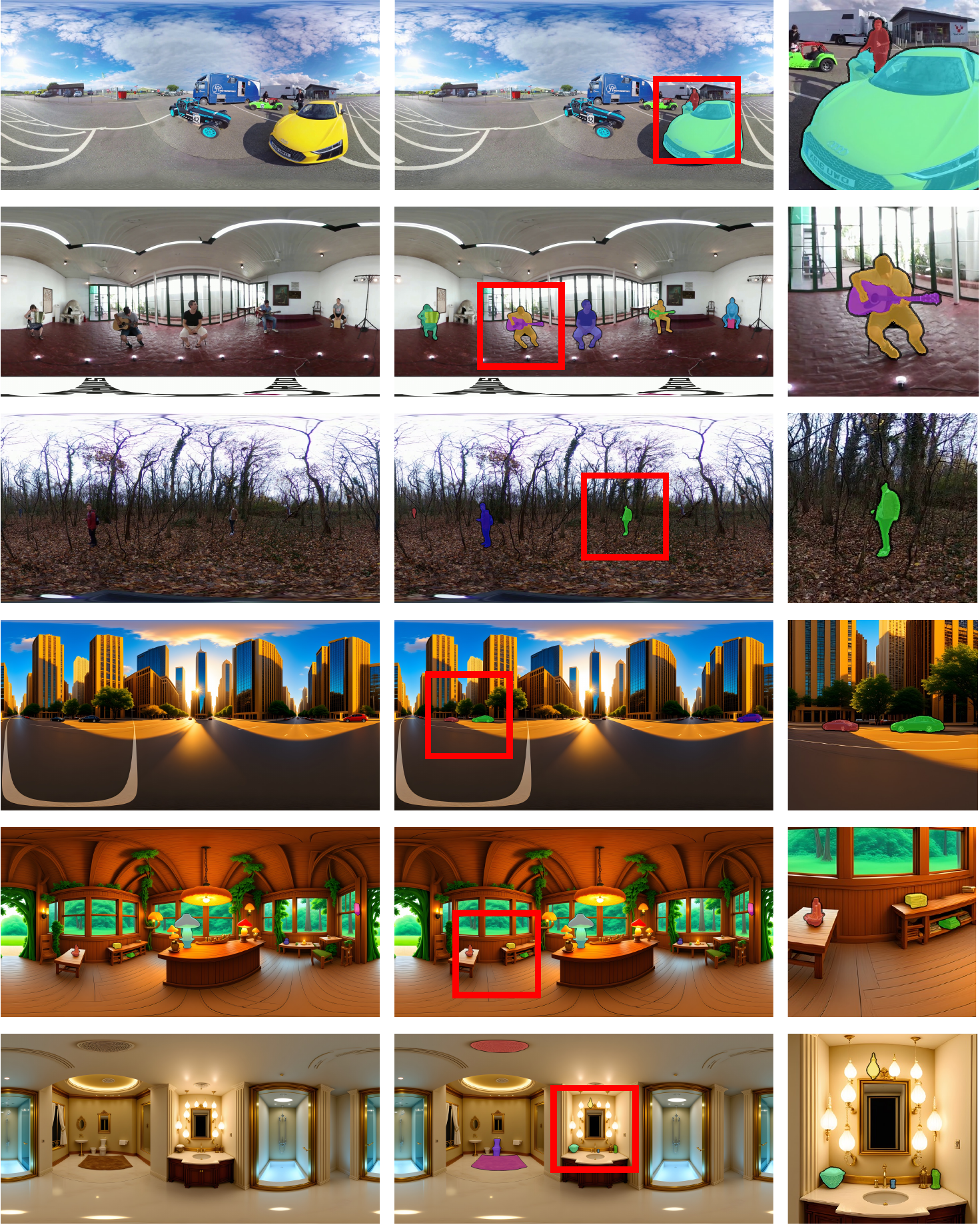}
    \caption{More segmentation results.}
    \label{fig:supple_qualitative}
\end{figure}

As shown in Fig.~\ref{fig:supple_qualitative}, we provide more segmentation visualization of our SAP.

